\pdfoutput=1
\documentclass[conference]{IEEEtran}
\IEEEoverridecommandlockouts
\usepackage{cite}
\usepackage{amsmath,amssymb,amsfonts}
\usepackage{algorithmic}
\usepackage{graphicx}
\usepackage{textcomp}
\usepackage{xcolor}
\usepackage[super]{nth}
\usepackage{setspace}
\usepackage{caption}
\usepackage[T1]{fontenc}
\usepackage{subfig}
\usepackage{multirow}
\usepackage{nth}
\usepackage{diagbox}
\usepackage{bm}

\def\BibTeX{{\rm B\kern-.05em{\sc i\kern-.025em b}\kern-.08em
    T\kern-.1667em\lower.7ex\hbox{E}\kern-.125emX}}

\begin{document}

\title{Blurring Fools the Network - Adversarial Attacks by Feature Peak Suppression and Gaussian Blurring}

\author{Chenchen Zhao and Hao Li$^{*}$
\thanks{This research work is supported by the SJTU (Shanghai Jiao Tong Univ.) Young Talent Funding (WF220426002).}
\thanks{Chenchen Zhao is with Dept. Automation, SJTU, Shanghai, 200240, China. }
\thanks{Hao Li, Assoc. Prof., is with Dept. Automation and SPEIT, SJTU, Shanghai, 200240, China. }
\thanks{* Corresponding author: Hao Li ({Email:\tt\small haoli@sjtu.edu.cn})}
}

\maketitle

\begin{abstract}
Existing pixel-level adversarial attacks on neural networks may be deficient in real scenarios,
since pixel-level changes on the data cannot be fully delivered to the neural network
after camera capture and multiple image preprocessing steps.
In contrast, in this paper, we argue from another perspective that gaussian blurring,
a common technique of image preprocessing, can be aggressive itself in specific occasions,
thus exposing the network to real-world adversarial attacks.
We first propose an adversarial attack demo named peak suppression (PS)
by suppressing the values of peak elements in the features of the data.
Based on the blurring spirit of PS, we further apply gaussian blurring to the data,
to investigate the potential influence and threats of gaussian blurring to performance of the network.
Experiment results show that PS and well-designed gaussian blurring can form adversarial attacks
that completely change classification results of a well-trained target network.
With the strong physical significance and wide applications of gaussian blurring,
the proposed approach will also be capable of conducting real world attacks.
\end{abstract}

\section{Introduction}\label{intro}

With the applications of deep learning and neural networks in core technology,
robustness and stability of their performance gradually attract much more attention.
Researches have proved that even well-trained neural networks may be vulnerable to adversarial attacks:
manipulated slight changes on the data resulting in misjudgement
\cite{pgd,competition,cw,tmim,perturbation,perturbation2,fgsm,ead,unrestricted,deepfool,jsma},
or generated meaningness data resulting in high-confidence recognition
by the target network \cite{nguyen2015deep,type1}.
The above two spirits of attacks are respectively named type II and type I attacks,
defined in \cite{type1} according to the characteristics of modifications on data.
These attack approaches have shown promising results and ability of
misleading the networks to wrong judgements with common image datasets (e.g. ImageNet \cite{imagenet}).

While the existing attack approaches are indeed threatening
the safety of neural networks and their applications,
their performance may be greatly reduced in real world scenarios.
In real image processing cases, there are several unique characteristics
of data and the processing system neglected by most attack approaches:
\begin{itemize}
\item
Reality of data.
The original data inputted to the network is collected from real scenes by non-ideal sensors,
which have internal distortion and slight errors on color capture.
Although existing type II adversarial attack methods are powerful
with their extremely small amplitude of data modifications,
it is barely possible for a real environment to correspond to the data with exact the same
distortion and color errors as the original data with the exact pixel modifications as expected.
Therefore, such adversarial examples do not have the same characteristic of reality
as that in the original data, and are barely possible to exist in the real world.
Reality loss also happens in type I attacks
with meaningless adversarial data \cite{nguyen2015deep} or manipulated data modifications \cite{type1}.
\item
Preprocessing.
The image data goes through several preprocessing steps such as
blurring, color transform, reshaping, etc. before inputted to the network.
Data modifications by a type II attack method may possibly fail after such preprocessing steps,
since the modification values are also changed in preprocessing together with the data,
with their influence possibly weakened or even reversed in the process.
\end{itemize}
The two points respectively correspond to the two perspectives of adversarial attacks:
whether the adversarial examples can truly exist in the real world;
whether the differences between the adversarial examples and the original data
can be fully delivered to the target network in real cases.
Negativity of either perspective results in deficiency of the attack method in real scenarios.
Unfortunately, the two points are just two of the main drawbacks of existing adversarial attack methods.

In this paper, in order to maintain the reality and aggressiveness of adversarial examples in real cases,
we choose not to make manipulated modifications on the data.
Instead, we focus on the image preprocessing step,
since it also makes changes to the raw data but does not affect the related data capturing process,
thus not affecting the reality of data.
Exploiting the vulnerability of the image preprocessing step is a better way
of generating real-world adversarial examples.
We demonstrate that gaussian blurring, a typical image preprocessing method,
is potentially threatening to neural networks itself by turning ordinary data aggressive
before inputting it to the network.
We first propose a novel adversarial attack demo named peak suppression (PS).
As a general type of blurring, PS suppresses the values of the peak elements (i.e. hotspots) in the features
and smoothes the features of data to confuse the feature processing module of the target network.
The feature-level blurring of PS generates adversarial examples
which are similar to the original examples after pixel-level blurring,
inspiring us to conduct adversarial attacks based on direct pixel-level blurring
happening in image preprocessing.
We further introduce gaussian blurring to adversarial attacks.
We construct several gaussian kernels that can change the data
into adversarial examples simply by gaussian blurring.
Such attacks may be more deadly, since a system with gaussian blurring as a part of preprocessing
may spontaneously introduce adversarial attacks to itself.
Our contributions are summarized as follows:
First, we propose a blurring demo to conduct several successful adversarial attacks;
second, we apply gaussian blurring to adversarial attacks,
and construct several threatening gaussian kernels for adversarial attacks;
last but not least, we prove that the commonly-used gaussian blurring technique in image processing
may be potentially `dangerous' and exposing the image processing network to adversarial attacks.
The attack approach proposed in this paper is effective and more importantly,
maintaining the reality of data and capable in real world scenarios,
since the designed gaussian kernels serve as a module in image preprocessing
and do not affect the original data.
This method raises a warning to image processing systems on their choice of parameters in gaussian blurring.

The paper is organized as follows:
Some related work of adversarial attacks is stated in Section \ref{relatedwork};
details of the peak suppression attack
and the gaussian blurring based attack are stated in Section \ref{method};
in Section \ref{experiment}, we conduct several experiments
to prove the effectiveness of PS and the gaussian blurring based attack;
we conclude this paper in Section \ref{conclusion}.

\section{Related Work}\label{relatedwork}

The gradient descent algorithm serves as the fundamental element
in most existing adversarial attack approaches.
Since the proposal of FGSM \cite{fgsm} stating the criterion of data modifications
to change the output of the network to the largest extent,
there are many studies aiming at modifying the data
to `guide' the network to a wrong result with gradient descent.
In \cite{deepfool}, L2 norm is used to determine the minimum value of perturbation in adversarial attacks;
authors in \cite{jsma} proposed a pixel-level attack method to minimize the number of changed pixels;
in \cite{pgd}, the authors used projection to avoid the situations
in which the modified data goes out of the data domain;
in \cite{perturbation, perturbation2}, the authors propose a feature-level attack method
by driving the intermediate feature output to that of another predefined data.
There are also defense-targeted attack methods (e.g. C\&W attack method \cite{cw} against distillation).

As stated in Section \ref{intro}, adversarial data generated by the attack methods above
may fail to maintain its reality.
Even the real-world attacks in \cite{real} have to go through a series of specific steps of data capture,
indicating that their performance relies heavily on the hardware environment and are partly unstable.
In this paper, apart from generating ordinary adversarial data with reality issues,
we also try to demonstrate that gaussian blurring, which is a common technique in image preprocessing,
has its potential threats to neural networks without affecting the reality of data.
With wrong parameter choices of gaussian blurring, an ordinary data sample collected from the real environment
can turn aggressive after preprocessing.

\section{The Proposed Gaussian Blurring Based Adversarial Attacks}\label{method}

\subsection{Feature peak suppression as demonstration}

We first propose an attack demo named Peak Suppression (PS) involving the basic spirit of blurring.

\begin{figure}[t]
\begin{center}
\includegraphics[width=0.95\linewidth]{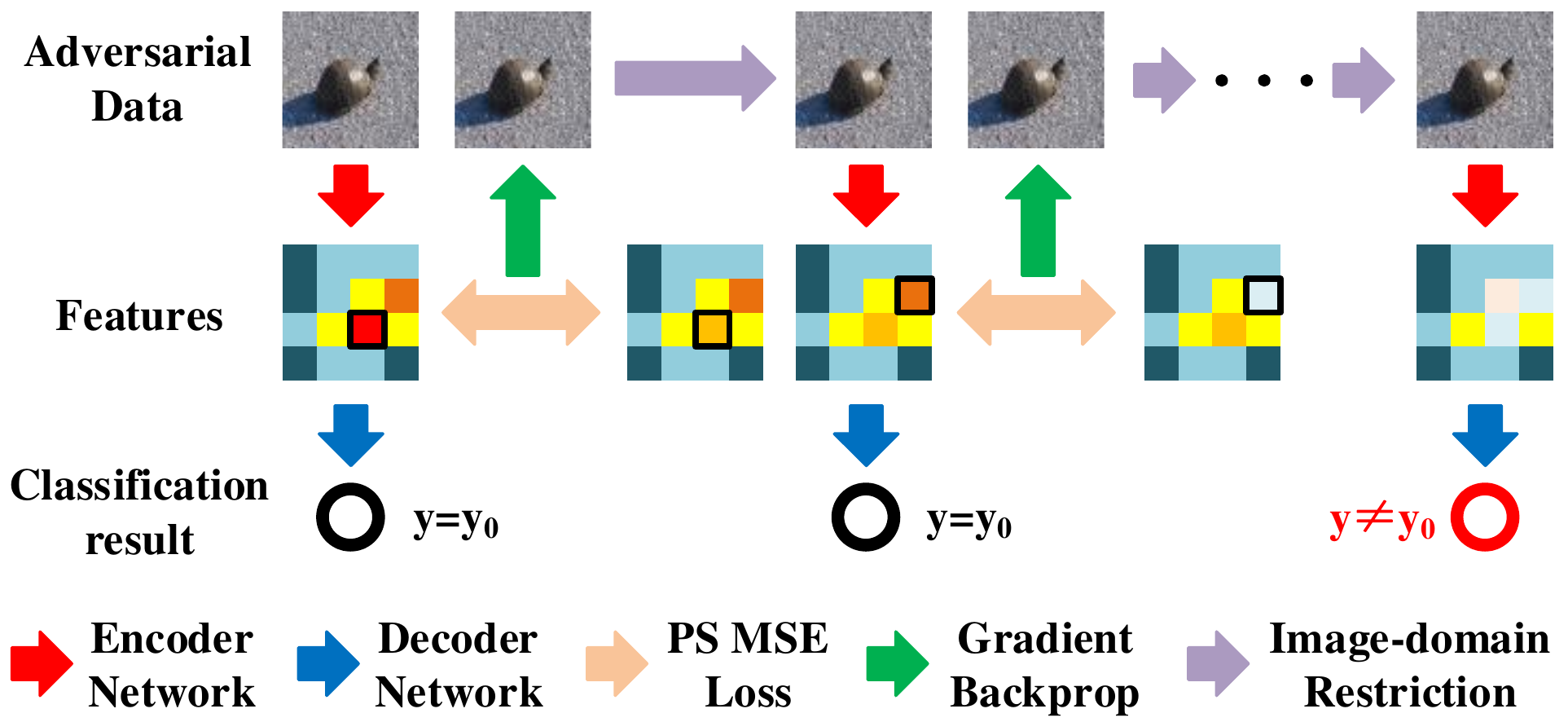}
\end{center}
\caption{The iteration process of Peak Suppression attack.}
\label{ps}
\end{figure}

As stated in \cite{cam}, CNN has relatively stronger responses in locations of key features of objects,
reflected by specific peak elements in feature outputs with much larger values.
Based on this, we try to confuse the network by suppressing such elements in the features,
to make the target network lose its concentration and make wrong judgements.
Then, adversarial data is derived from the peak-suppressed features.
The iteration process of PS is shown in Figure~\ref{ps}.
After finite number of iterations, peaks in the features turn implicit enough to confuse the network,
and the iteration process immediately ends when the network changes its judgement.
PS can be considered as a simple feature-level blurring technique.

In generation of the adversarial data $x$, the criterion is shown in \eqref{ps-eq},
in which $f_{encoder}(x)_i$ is the $i$th layer of the feature maps.
\begin{equation}
\begin{aligned}
L_{PS} = \sum_i ||max(f_{encoder}(x)_i)-\overline{f_{encoder}(x)_i}||_2
\end{aligned}
\label{ps-eq}
\end{equation}
PS is a white-box non-targeted attack approach with this criterion.

After every iteration, we apply the image-domain restrictions to the generated data,
with its form shown in \eqref{pgd}.
\begin{equation}
\begin{aligned}
x = min(max(x, \bm{0}), \bm{1})
\end{aligned}
\label{pgd}
\end{equation}

\subsection{Gaussian blurring based adversarial attack}

Proved in experiments in Section \ref{experiment}, with features smoothed,
the adversarial data generated by PS also has lower clarity than the corresponding original data,
and is similar to the results by pixel-level blurring.
Therefore, although PS is not an image preprocessing method,
it inspires us to investigate and conduct experiments
on pixel-level blurring based adversarial attacks happening in image preprocessing.
We further propose a novel adversarial attack method based on the typical gaussian blurring technique.
Instead of determining specific pixels or patches to modify,
we directly apply gaussian blurring to the original data and try to turn it aggressive.
With the iterations, the gaussian kernel is modified instead of the original data.
One main difference between blurring in PS and gaussian blurring based attack
is that the features are blurred in PS,
while the original data is blurred in the gaussian blurring based attack.
The latter attack method better fits the real cases
in which blurring is applied to the data instead of its features.

\begin{figure}[t]
\centering
\includegraphics[width=0.95\linewidth]{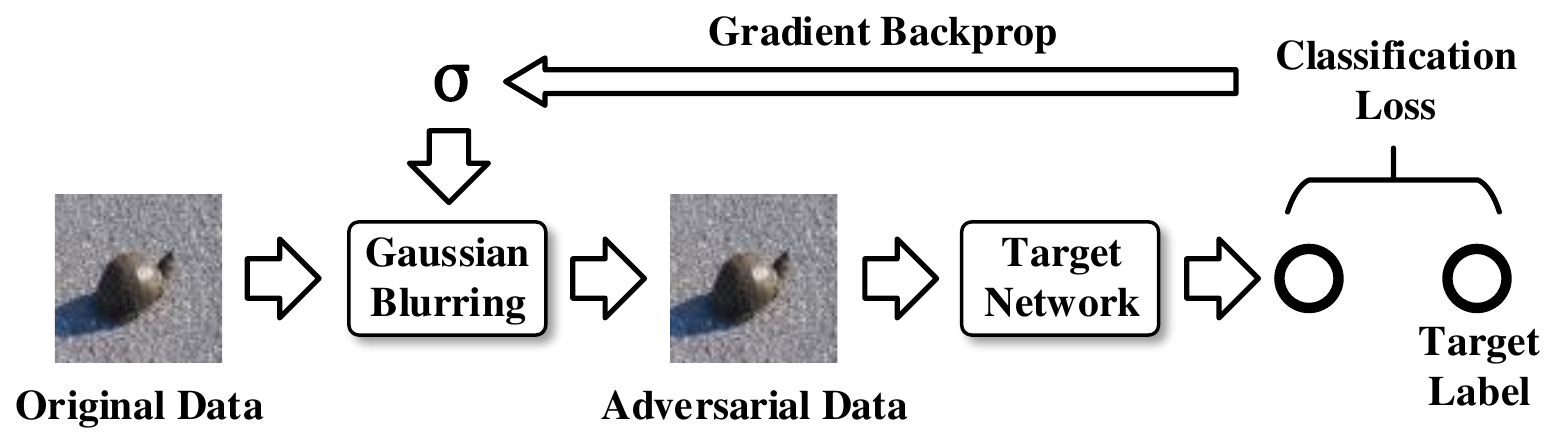}
\caption{Structure of the gaussian blurring based attack.}
\label{structure}
\end{figure}

Structure of the proposed method is shown in Figure~\ref{structure}.
In every iteration, the original data is blurred with a designed kernel
and inputted to the target network afterwards.
The designed kernel has the same number of channels as the data (i.e. 3)
with different channel having different std values.
The attack method adopts the originally adopted criterion (by the target network),
but with the iteration direction just the opposite, shown in \eqref{g-eq},
in which $k$ is the gaussian kernel with mean $\mu$ and std $\sigma$.
Note that the criterion of the gaussian blurring based attack can be set targeted,
by changing $y_0$ to a predefined label and reversing the iteration direction.
However, due to the strong constraint of data modifications (to satisfy the equation of gaussian blurring),
we do not expect the adversarial data to be classified as a certain class.
Instead, we set the criterion non-targeted, which are proved effective in Section \ref{experiment}.
\begin{equation}
\begin{aligned}
L_g = -L_0(f(g(x, k(\mu, \sigma))), y_0)
\end{aligned}
\label{g-eq}
\end{equation}
When back propagating, $\mu$ of the gaussian kernel is fixed as the center of the kernel;
$\sigma$ is a trainable parameter with gradient determined in \eqref{grad},
in which $(\mu_1, \mu_2)$ is the coordinate of the center of the kernel;
$(\sigma_1, \sigma_2)$ is the std value of the 2D gaussian distribution
corresponding to one channel of the kernel;
$N$ is the predefined scale of the gaussian kernel.
\begin{equation}
\begin{aligned}
\frac{\mathrm{d}L_g}{\mathrm{d}\sigma_1} &= \frac{\mathrm{d}L_g}{\mathrm{d}k(\mu, \sigma)}\cdot\frac{1}{N^2}\sum_{(a,b)\in k}\frac{((a-\mu_1)^2 - \sigma_1^2)\cdot A}{2\pi\sigma_1^4\sigma_2} \\
\frac{\mathrm{d}L_g}{\mathrm{d}\sigma_2} &= \frac{\mathrm{d}L_g}{\mathrm{d}k(\mu, \sigma)}\cdot\frac{1}{N^2}\sum_{(a,b)\in k}\frac{((b-\mu_2)^2 - \sigma_2^2)\cdot A}{2\pi\sigma_1\sigma_2^4} \\
A &= e^{-\frac{1}{2}(\frac{(x - \mu_1)^2}{\sigma_1^2} + \frac{(y - \mu_2)^2}{\sigma_2^2})}
\end{aligned}
\label{grad}
\end{equation}
The iteration process immediately ends when the network changes its judgement.

With the designed gaussian blurring, the modified data will constantly be in the image domain $[0,1]_n$,
and it is possible to generate adversarial data free from the out-manifold-violation raised in \cite{type1}.

\begin{figure}
\centering
\includegraphics[width=0.95\linewidth]{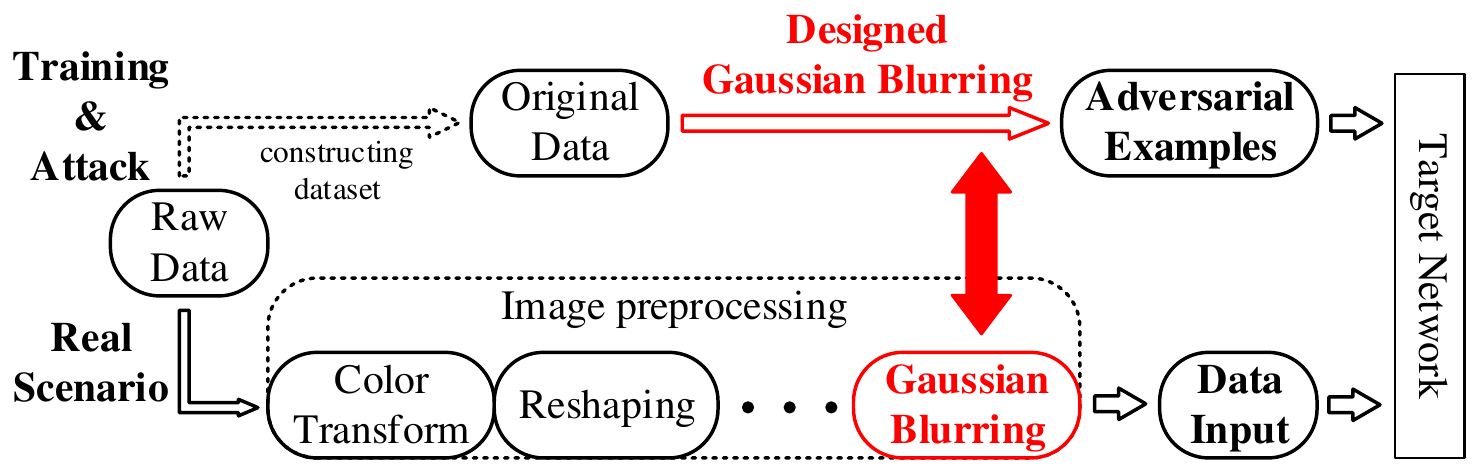}
\caption{The gaussian blurring based attack serves as a warning to existing image processing system with gaussian blurring.}
\label{system}
\end{figure}

With the wide adoption of gaussian blurring in image processing systems,
the designed gaussian kernels are highly possible to exist in real cases,
and generate real-world adversarial examples via gaussian blurring preprocessing.
Note that the parameters of gaussian blurring in the target image processing systems
are unknown and unchangable, so the designed aggressive gaussian blurring
is more like a warning to existing systems of this kind
(shown in Figure~\ref{system}) instead of a pure attack method.
It raises a fact that wrong parameter choice of gaussian blurring in image processing systems
may produce a terrible preprocessing module that turns ordinary data into adversarial examples.

\section{Experiments and Discussions}\label{experiment}

\subsection{Experiment setup}

Several PyTorch-pretrained classic neural network models:
ResNet18, ResNet50 \cite{resnet}, DenseNet121 \cite{densenet}
with ImageNet \cite{imagenet} are used as the target networks
for adversarial data generation, baseline comparisons and investigating the relationship
between the scale of the gaussian kernel, initial $\sigma$ value and attack performance
in the gaussian blurring based attack.
PS and the gaussian blurring based attack are conducted with the same ImageNet dataset.
Detailedly, the dataset involves five classes (throne-857, tractor-866, mushroom-947,
turtle-34 and breastplate-461) with 50 images in each.
Each image in PS and gaussian kernel in the gaussian blurring based attack
goes through $5 \times 10^3$ iterations with step length 1.
Choice of the relatively large step length in PS is because
step length 1 meets the requirements of setting the hotspots to average;
this choice in the gaussian blurring based attack is because
gaussian blurring does not have the ability of changing characteristics of the data
(such characteristics do \textbf{NOT} refer to the features outputted by the network).
Note that with all networks, PS is applied to the deepest features of the data.

Several attack baselines are experimented to prove the effectiveness of the proposed methods.
The deepest features of the adversarial data generated by PS and the gaussian blurring based attack
during the iterations are also visualized and analyzed.

\subsection{Attack results of PS}

\begin{figure}[t]
\centering
\includegraphics[width=0.95\linewidth]{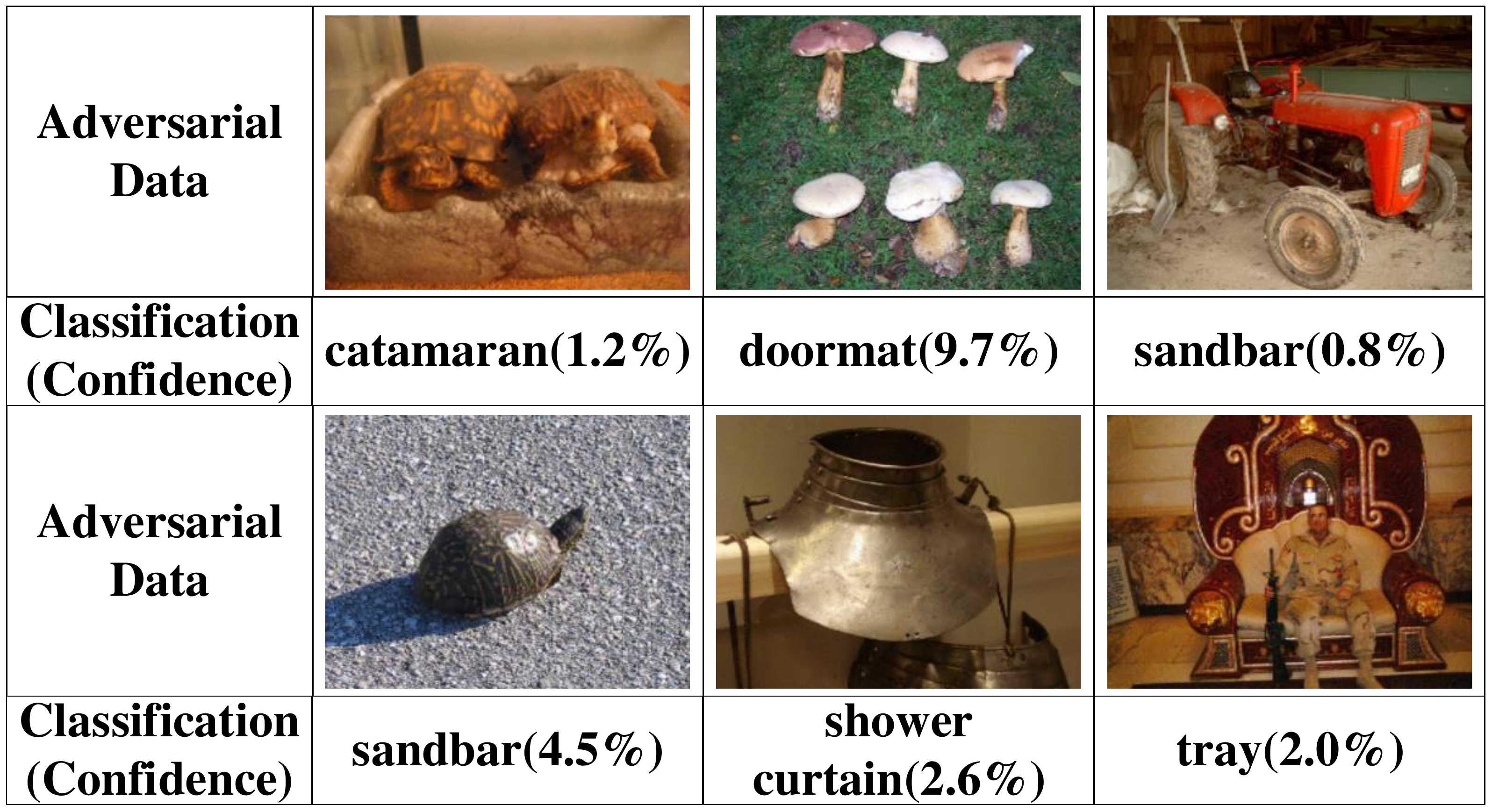}
\caption{Adversarial examples generated by PS.}
\label{psresult}
\end{figure}

\begin{figure}[t]
\centering
\includegraphics[width=0.7\linewidth]{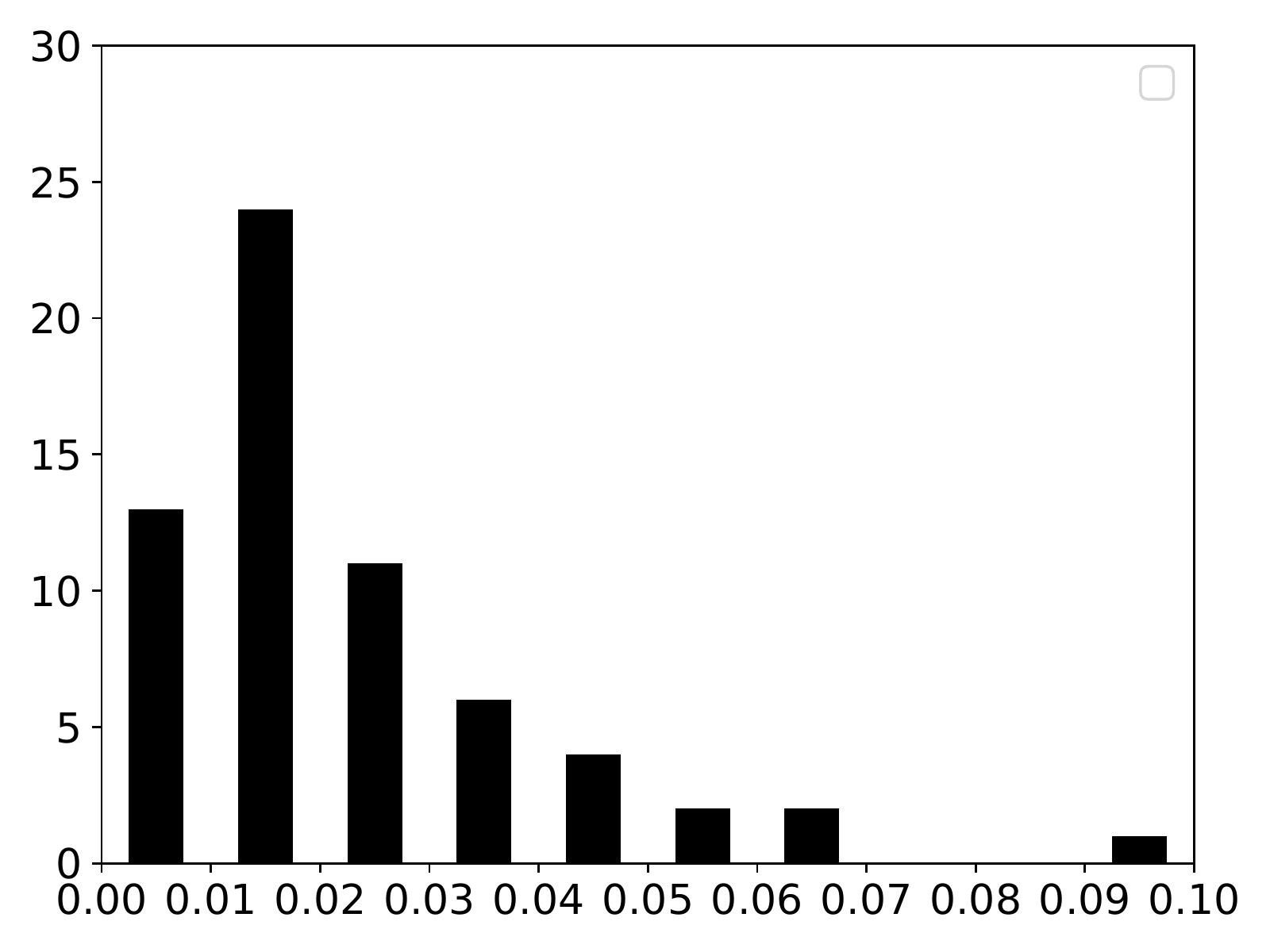}
\caption{The confidence distribution of the low-confidence adversarial examples by PS.}
\label{psconfidence}
\end{figure}

PS achieves good performance of adversarial attack,
with the generated adversarial examples (with ResNet18) shown in Figure~\ref{psresult}.
From Figure~\ref{psresult}, most adversarial examples have quite low misclassification confidence.
In the experiment, 85.3\% of the adversarial examples have the misclassification confidence lower than 0.1.
with the confidence distribution of such examples shown in Figure~\ref{psconfidence}.
Two main reasons lead to such low confidence:
First, PS adopts a non-targeted criterion,
and it is barely possible for it to generate an example with extremely high confidence;
second, PS suppresses the hotspots in the features,
indicating that the examples PS generates do not have obvious characteristics of any classes.

\begin{figure}[t]
\centering
\subfloat[The original data.]{
\includegraphics[width=0.35\linewidth]{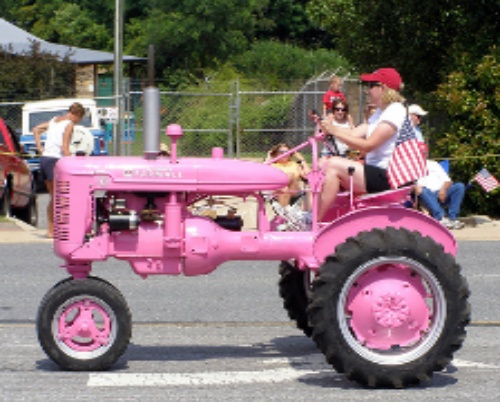}
}

\subfloat[The confidence decrease.]{
\includegraphics[width=0.45\linewidth]{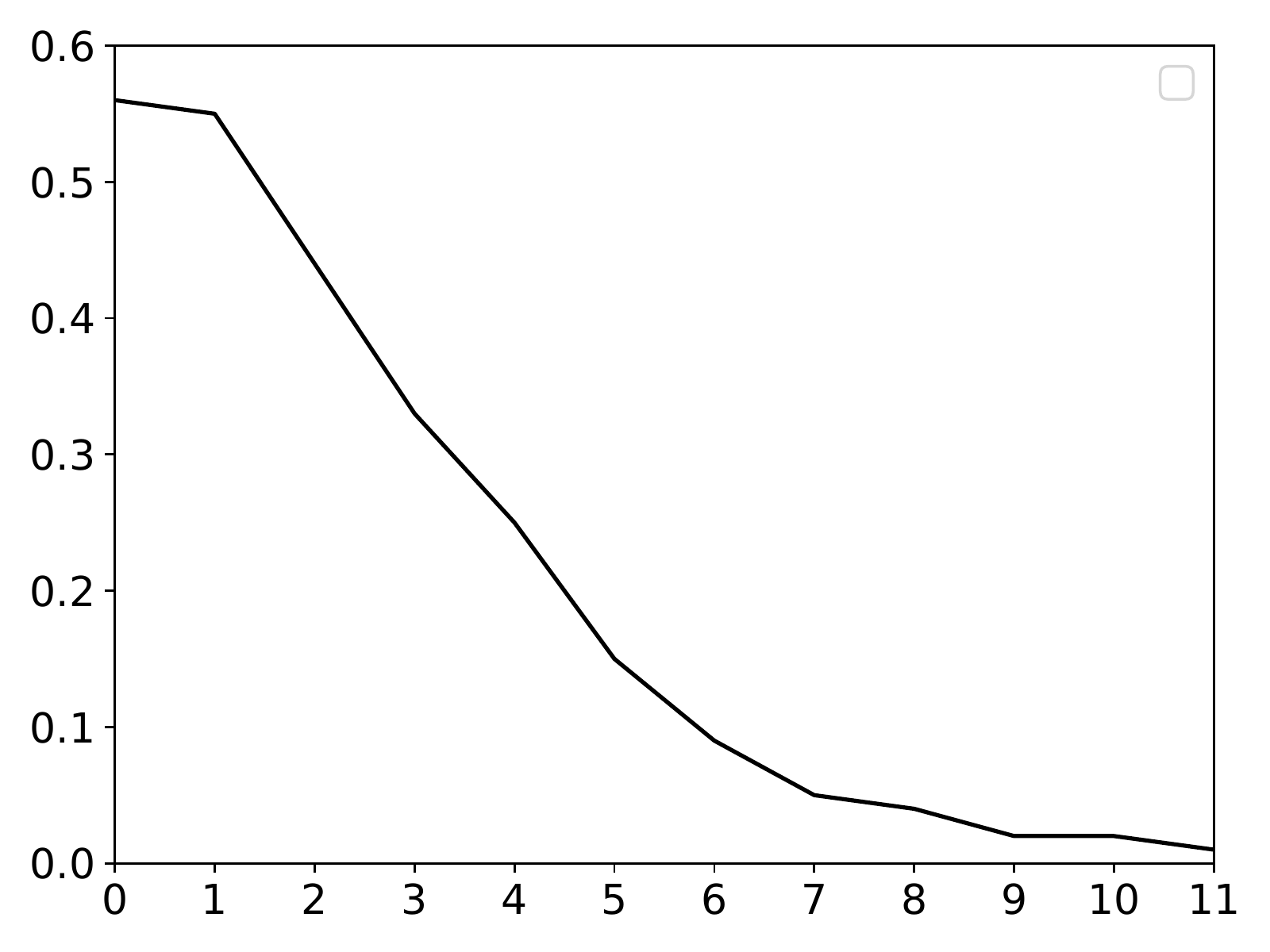}
\label{ps_progress_confidence}
}
\subfloat[The max value decrease.]{
\includegraphics[width=0.45\linewidth]{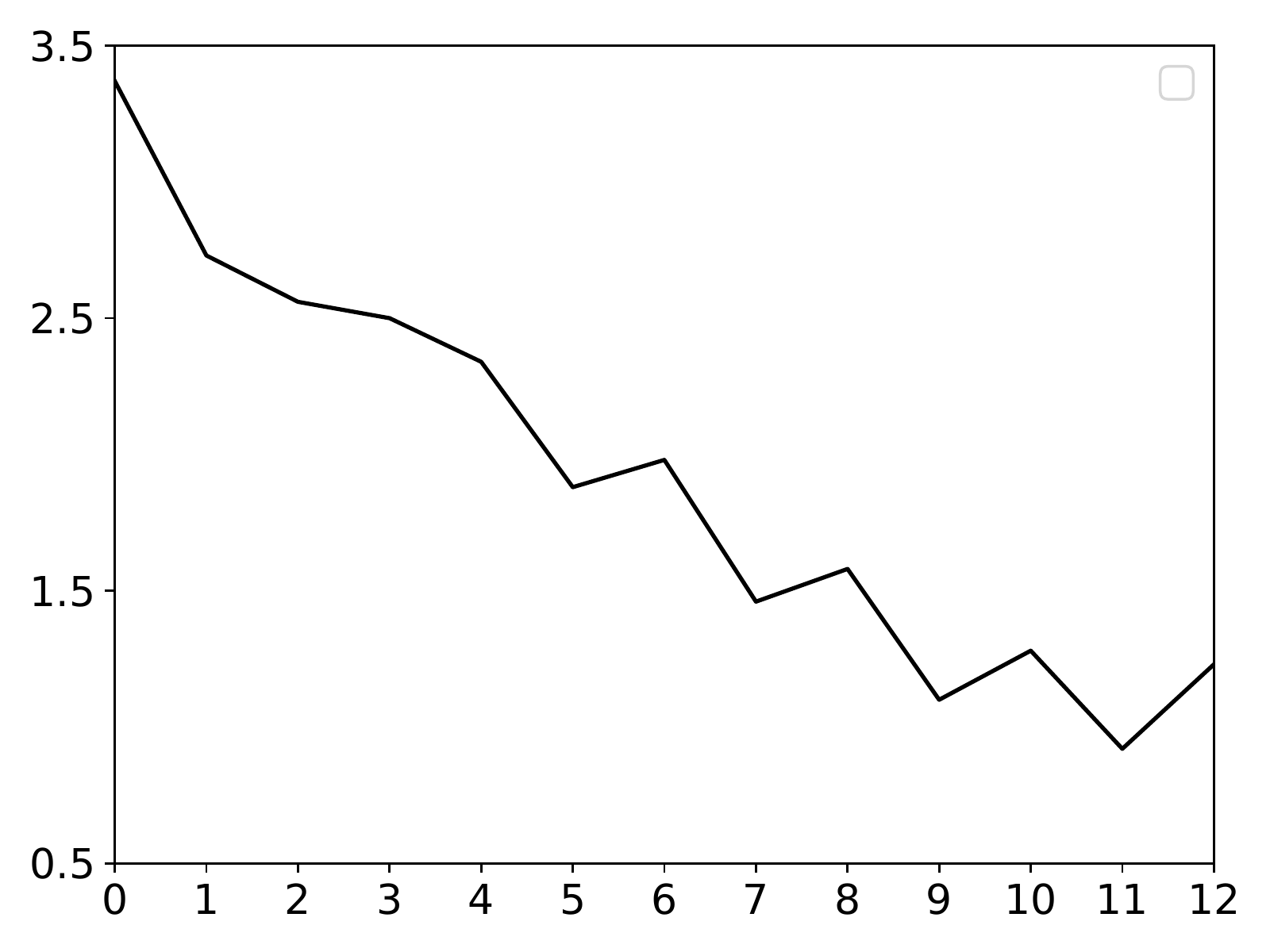}
\label{ps_progress_max}
}

\subfloat[The feature heatmap.]{
\includegraphics[width=0.95\linewidth]{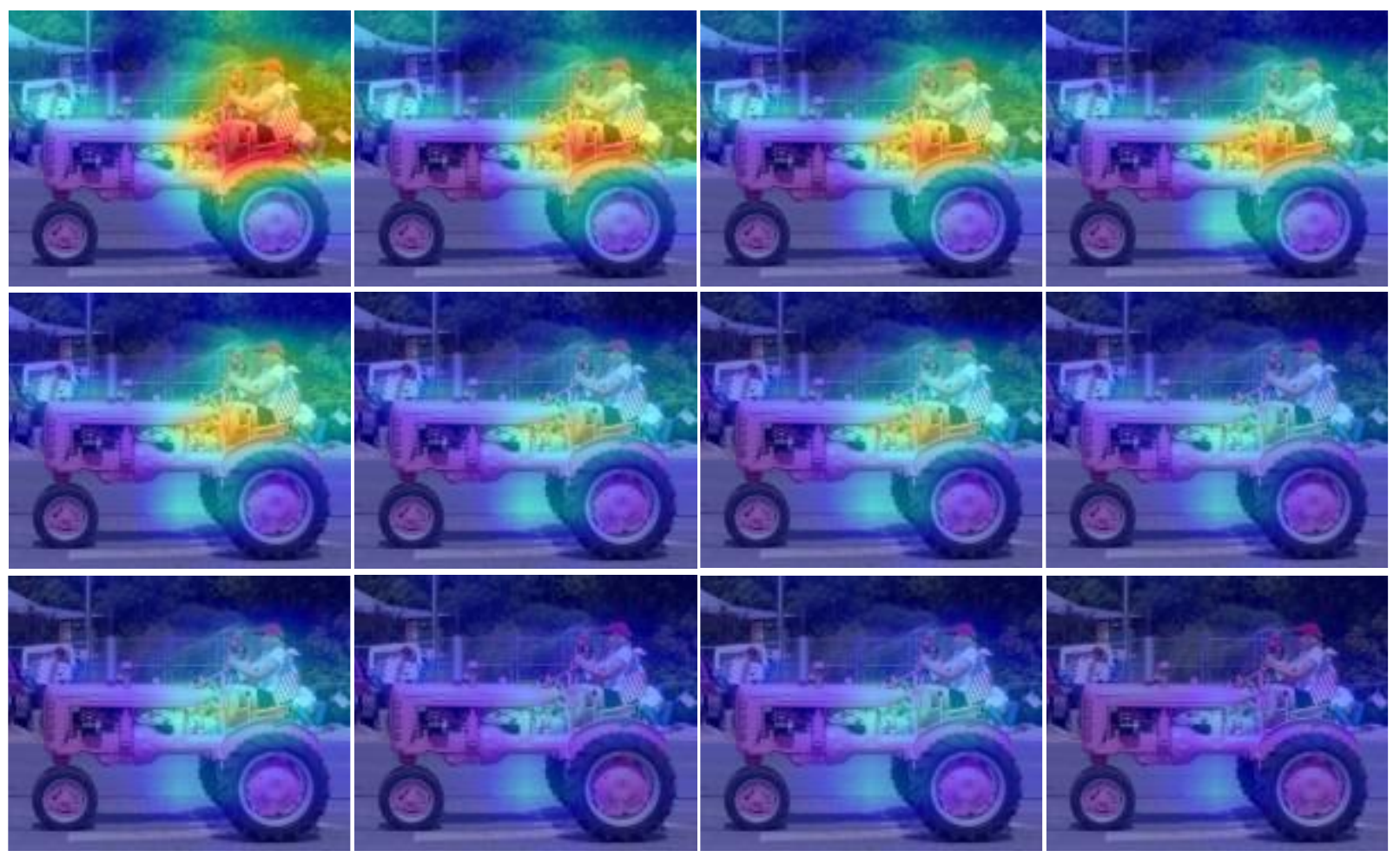}
\label{ps_progress_visual}
}
\caption{The visualization process of PS.}
\label{ps_progress}
\end{figure}

Figure~\ref{ps_progress} shows the visualized iteration process of PS on an image.
The data goes through 12 iterations and changes the classification result of the network
from tractor (866) to manhole cover (640).
Figure~\ref{ps_progress_visual} shows heatmaps of a randomly-selected channel
of the deepest features in the iteration process;
Figure~\ref{ps_progress_confidence} shows the confidence decrease of class 866;
Figure~\ref{ps_progress_max} shows the decrease of the maximum value of the feature.
The visualization verifies the feature-peak-suppression aim of PS.

\subsection{Attack results of the gaussian blurring based attack}

$\sigma$ is set 10 initially in the gaussian blurring based attack.
Choice of the relatively large initialization value of $\sigma$ is further explained in Section \ref{theory}.
It is found that a gaussian kernel with $\sigma=10$ does not result in qualitative changes of the data,
in which qualitative changes are defined by `change of judgement by the oracle' \cite{type1}.

\begin{figure}[t]
\centering
\includegraphics[width=0.95\linewidth]{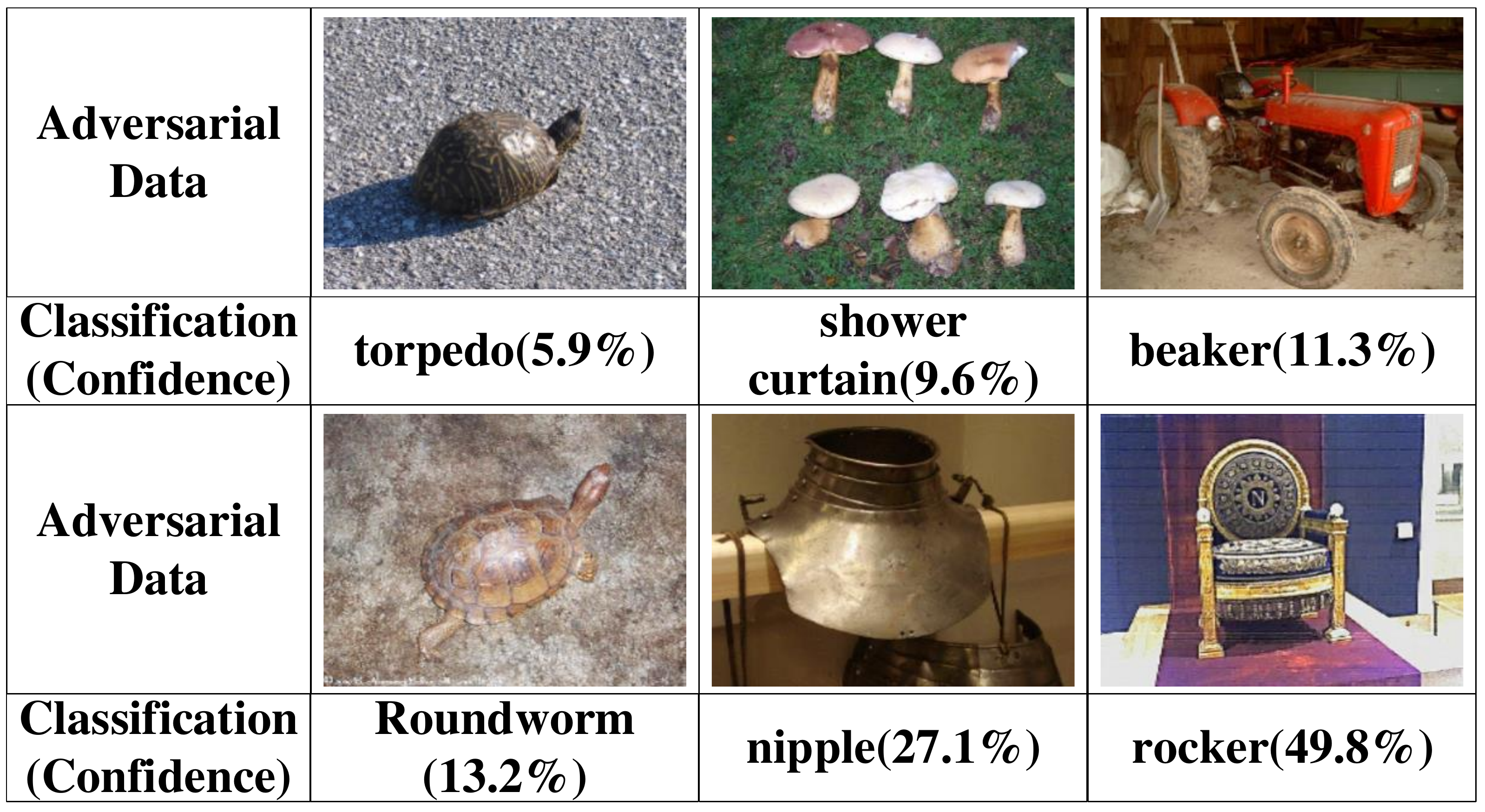}
\caption{Adversarial examples generated by the gaussian blurring based attack.}
\label{gresult}
\end{figure}

The gaussian blurring based attack is also effective,
with the generated adversarial examples (with ResNet18) shown in Figure~\ref{gresult}.
Similar to PS, adversarial examples generated by the gaussian blurring based attack
also have relatively low misclassification confidence,
which is also closely related to the non-targeted criterion it adopts.

\begin{figure}
\centering
\includegraphics[width=0.95\linewidth]{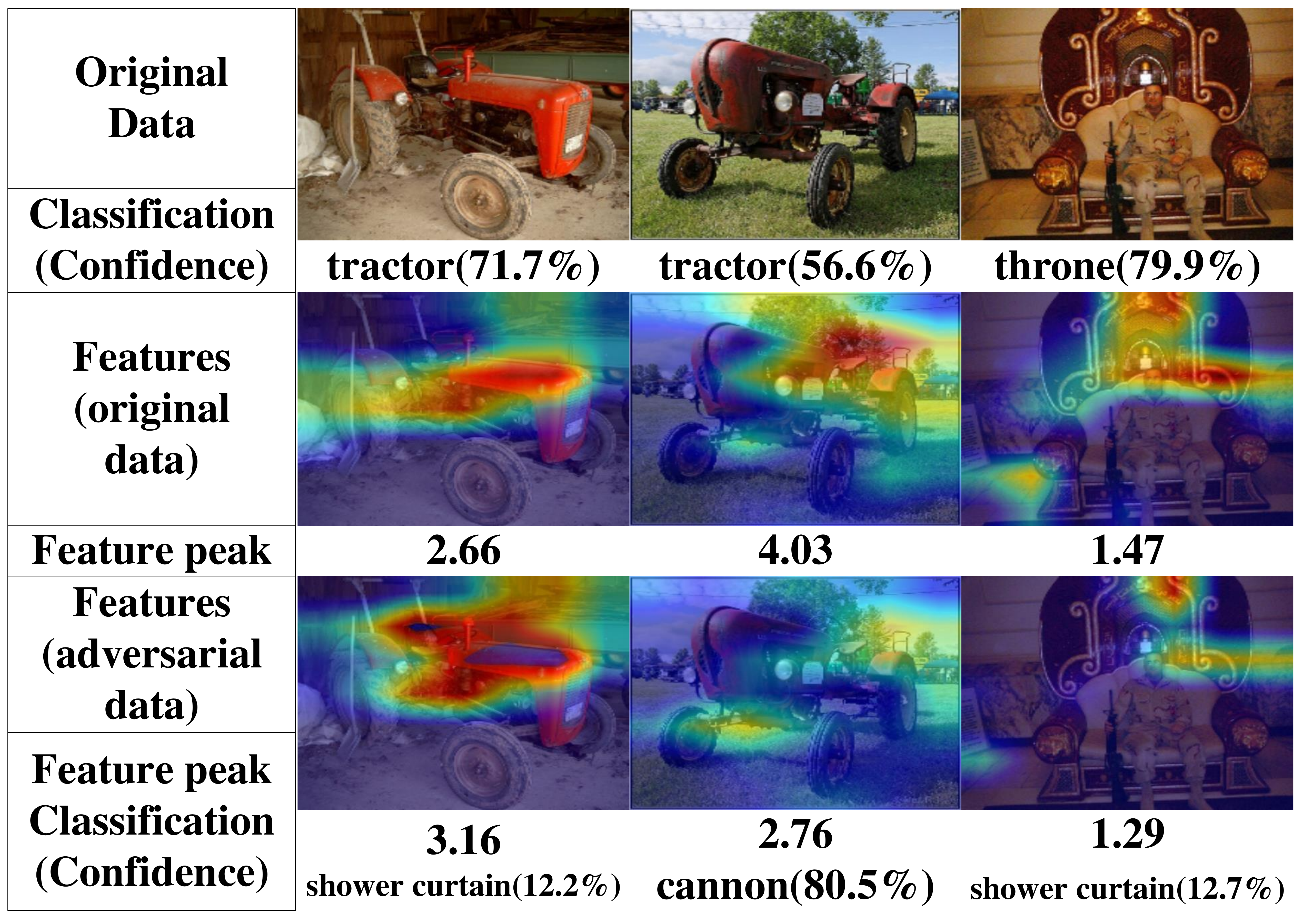}
\caption{Visualization of the features of the adversarial data generated by the gaussian blurring based attack.}
\label{gaussianvis}
\end{figure}

A similar visualization is conducted with the gaussian blurring based attack.
Multiple data samples are converted to adversarial examples with related features visualized.
The visualization results are shown in Figure \ref{gaussianvis}.
From the results, although gaussian blurring does not suppress the feature peak values as PS does,
it still changes the attention of the target network and its response to key features
(shown in Figure \ref{gaussianvis} that hotspots (red) turn `cold' (blue) sharply).
Compared with PS, the gaussian blurring based attack is more like a non-targeted feature peak shifting
rather than feature peak suppression.
Note that in Figure \ref{gaussianvis}, the adversarial data is generated with only 
one single gaussian blurring step with kernel scale 9 and $\sigma=10$,
and without iterations of $\sigma$.
In the experiment, 97.1\% (ResNet18), 98.1\% (ResNet50) and 97.6\% (DenseNet121)
of the original data turns aggressive with only one single step.
Reasons of this single-step iteration phenomenon are stated and analyzed in Section \ref{theory}.

We do not investigate the changing process of $\sigma$ of the gaussian kernel.
Reasons are stated in Section \ref{theory}.

\subsection{Attack performance with different kernel scales in the gaussian blurring based attack}\label{differentscale}

\begin{figure}[t]
\centering
\includegraphics[width=0.7\linewidth]{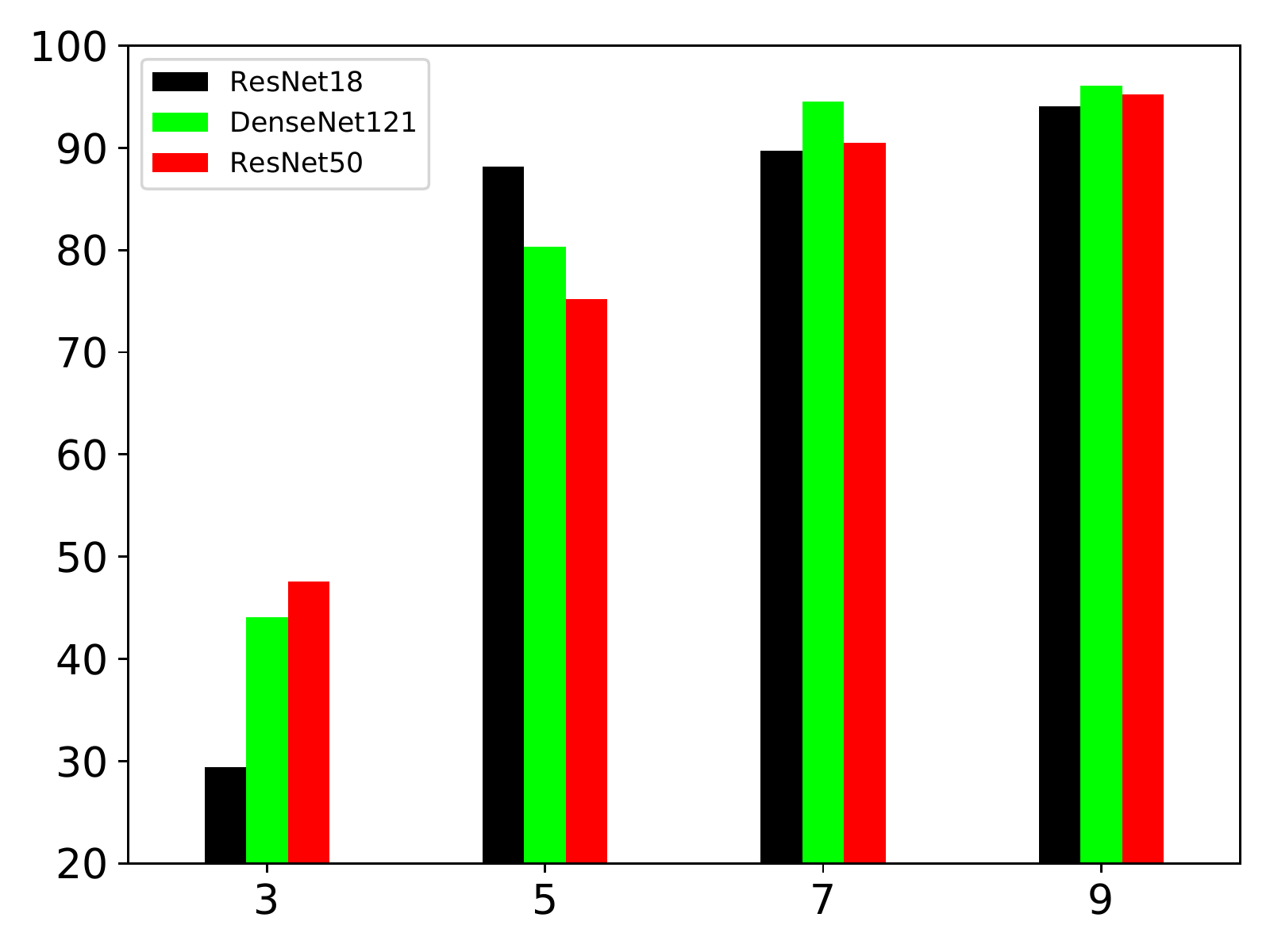}
\caption{Error rates of the three models caused by different kernel scale in the gaussian blurring based attack.}
\label{scale}
\end{figure}

Different scales of gaussian kernels result in different attack performance.
By setting the blurring scales to 3, 5, 7 and 9,
the gaussian blurring based attack on the three models have the error rates shown in Figure~\ref{scale}.
With the scale turning larger, the data goes through greater modifications,
and is more likely to be misclassified.
In addition, it is also found that no kernel scale result in qualitative changes of the data.

\subsection{Baseline comparisons}

We conduct baseline comparisons with several existing attack methods.
We conduct the same experiments as in \cite{perturbation2} but without transferability tests,
since both PS and the gaussian blurring based attack are non-targeted,
and cannot be measured by targeted success rate raised in \cite{perturbation2}.
Attack baselines involve targeted projected gradient descent (tPGD) \cite{pgd},
targeted momentum iterative method (tMIM) \cite{tmim}
and feature distribution attack (FDA) \cite{perturbation}.
The gaussian blurring based attacks with kernel scale 3 and 9 are involved.

Error rates of the DenseNet121 and ResNet50 models
with the five attack methods are shown in Table \ref{attacktable}.
From the comparison results, PS and the gaussian blurring based attack with kernel scale 9
reach the state-of-the-art attack performance.
In comparison, kernel scale 3 is rather deficient,
which is similar to the results shown in Section \ref{differentscale}.
The gaussian blurring based attack is slightly worse than PS,
since gaussian blurring greatly restricts the range of modifications on data.

\begin{table}[]
\centering
\begin{tabular}{|c|c|c|c|}
\hline
 & tPGD & tMIM & FDA \\
\hline\hline
DenseNet121 & 23.1 & 48.6 & 91.9 \\
ResNet50 & 20.2 & 44.3 & 87.3 \\
\hline
 & PS(ours) & G-3(ours) & G-9(ours) \\
\hline\hline
DenseNet121 & 97.6 & 44.1 & 96.1 \\
ResNet50 & 99.0 & 47.6 & 95.2 \\
\hline
\end{tabular}
\caption{Baseline comparisons on error rates.}
\label{attacktable}
\end{table}

The table serves as a reference only, which is because of two main differences
between our proposed methods and these baseline methods:
first, our methods do not consider transferability as a factor of performance;
second, since blurring is an intermediate step with continuous-domain inputs,
our methods do not consider the discreteness of the image domain.

According to the numerical results and visualizations above,
we may reach the following conclusions of the relationship between blurring and adversarial attacks:
\begin{itemize}
\item
A designed feature-level blurring can mislead the network to wrong judgements,
since the attention of the target network can be distracted.
\item
A feature-level blurring may be equivalent to a specific pixel-level blurring.
However, a pixel-level blurring may not be equivalent to any feature-level blurring.
\item
Adversarial examples have different feature peaks from the corresponding original data,
and can be generated by multiple feature-manipulation methods
not limited to peak suppression and peak shifting.
\end{itemize}

\subsection{Theoretical analyses and discussions - A manifold perspective}\label{theory}

For clearer illustration, we use the two-nested-spirals model as an example
to investigate the essential factors to make the gaussian blurring based attack work.

\begin{figure}[t]
\centering
\subfloat[An MLP classifying points on a two nested spiral pair.]{
\includegraphics[width=0.45\linewidth]{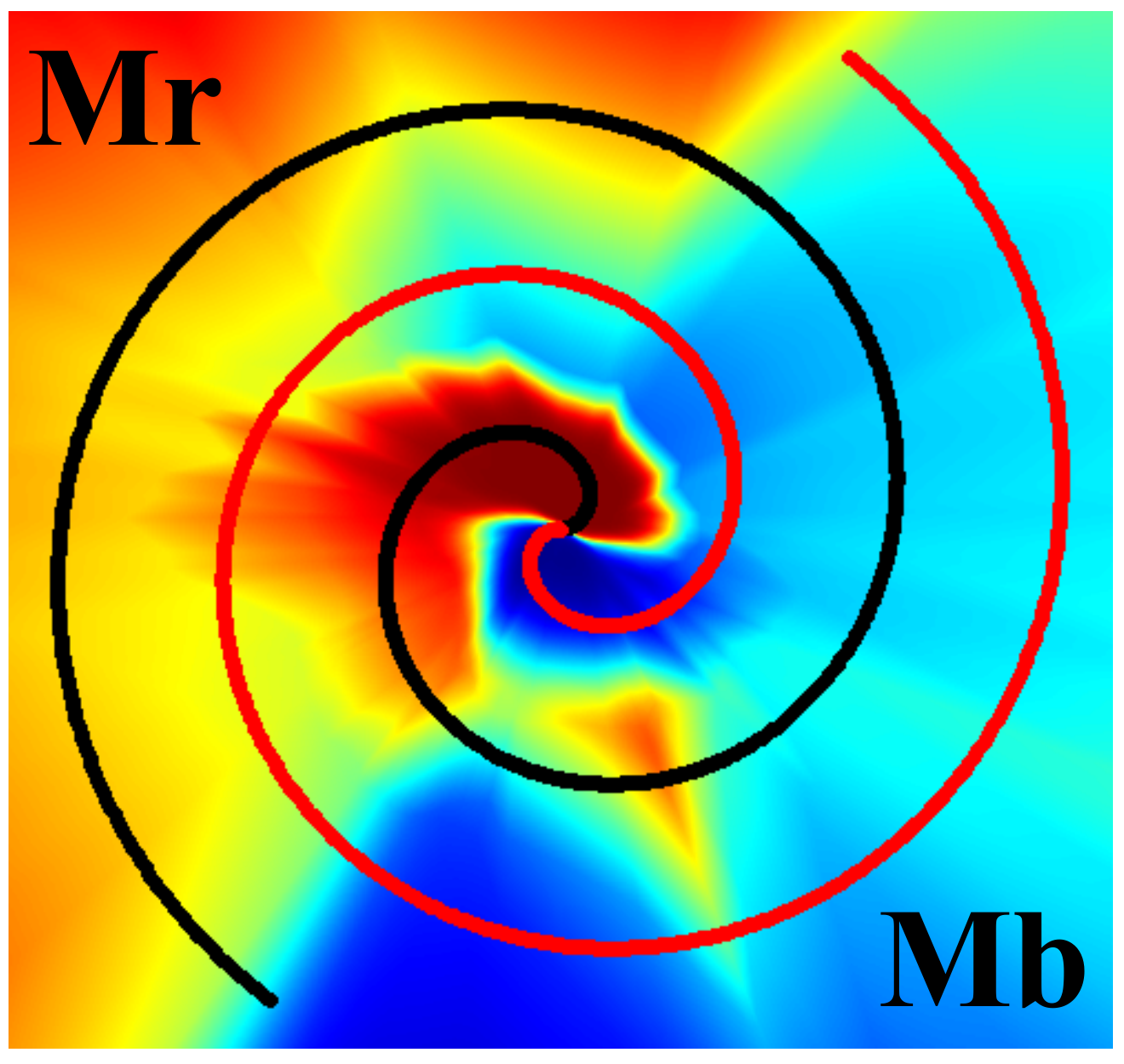}
\label{spirals}
}
\subfloat[The manifold of a data point after different gaussian blurring operations.]{
\includegraphics[width=0.45\linewidth]{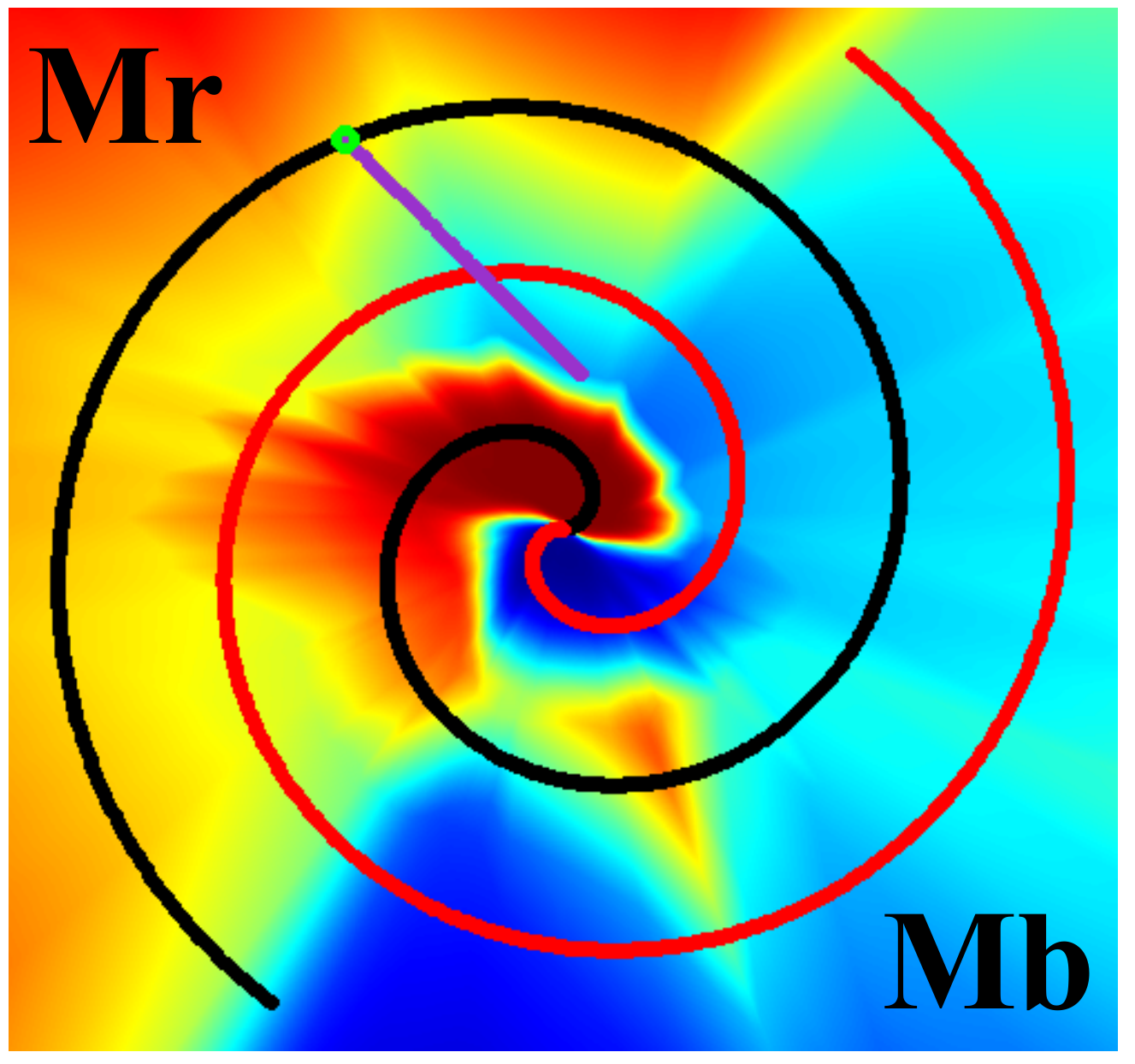}
\label{gaussianline}
}
\caption{A two-nested-spirals model for illustration.}
\end{figure}

Figure~\ref{spirals} shows a pair of two nested spirals.
A 5-layer multi-layer perceptron (MLP) is trained to classify points on the two spirals.
The red and blue areas are 2D manifolds representing the data domain of two classes
in the perspective of the MLP, named manifold $M_r$ and $M_b$.
Since the problem is much easier compared with the ImageNet classification in Section \ref{experiment},
training of the MLP ends when it reaches 70\% accuracy,
to represent that networks are not that accurate when facing harder problems.

When applying gaussian blurring to a specific point on the spirals,
the modified data points form a manifold $M$ with $\sigma$ varying in $(0, \infty)$.
According to characteristics of convolution, the data before and after gaussian blurring
(respectively $D_0$ and $D$) satisfy the following condition:
\begin{equation}
\begin{aligned}
&D = D_0 \otimes K(\mu,\sigma) \\
&\to \sum_{d \in D}d = \sum_{d_0 \in D_0}d_0 \times \sum_{k \in K}k = \sum_{d_0 \in D_0}d_0 \\
\end{aligned}
\label{eqconv}
\end{equation}
Therefore, the sum of elements of the data before and after gaussian blurring remains the same.
This indicates that the manifold $M$ is a curve within a specific high-dimensional plane,
shown in \eqref{eqplane}.
\begin{equation}
\begin{aligned}
M \subset \{p|\sum_{p_i \in p}p_i=\sum_{d_0 \in D_0}d_0\}
\end{aligned}
\label{eqplane}
\end{equation}
The two endpoints of the curve are the original data point
and the modified data after gaussian blurring with $\sigma \to \infty$.
For a 1D gaussian blurring on the original data $D_0$ (shown in Figure~\ref{gaussianline} as the green point),
$M$ is a straight line segment (shown in Figure~\ref{gaussianline} as the purple line).

From Figure~\ref{gaussianline}, $M$ spans both $M_r$ and $M_b$.
For any $D_0$, if the corresponding manifold $M$ spans both $M_r$ and $M_b$,
adversarial data points can be generated by `dragging' the original data point to the other manifold,
with the mathematical representation shown in \eqref{math}.
\begin{equation}
\begin{aligned}
&M_r \cap M \neq \phi,M_b \cap M \neq \phi, \\
&D_0 \in M_0 (M_0 \in \{M_r,M_b\}) \\
&\to \exists \sigma : D_0 \otimes K(\mu,\sigma)\in \{M_r,M_b\}^{M_0} \\
\end{aligned}
\label{math}
\end{equation}
However, such `adversarial' data may essentially belong to a class different from $D_0$,
or even not belong to any classes
(e.g. in this illustration that the other endpoint of the line segment belongs to neither of the classes).
Typical instances of such data are the meaningless examples
generated by type I attack methods \cite{nguyen2015deep}.

In real scenarios, however, we can make a reasonable assumption that
gaussian blurring with a moderate $\sigma$ does not qualitatively change the data:
\begin{equation}
\begin{aligned}
&\forall D,\exists \sigma_0: \\
&\forall \sigma \in (0,\sigma_0), Oracle(D) = Oracle(D\otimes K(\mu,\sigma))
\end{aligned}
\label{eqsigma}
\end{equation}
Therefore, in real cases, all data points in $M$ belong to the same class,
and the adversarial data generated in \eqref{math} turns effective.

\begin{figure}[t]
\centering
\includegraphics[width=0.95\linewidth]{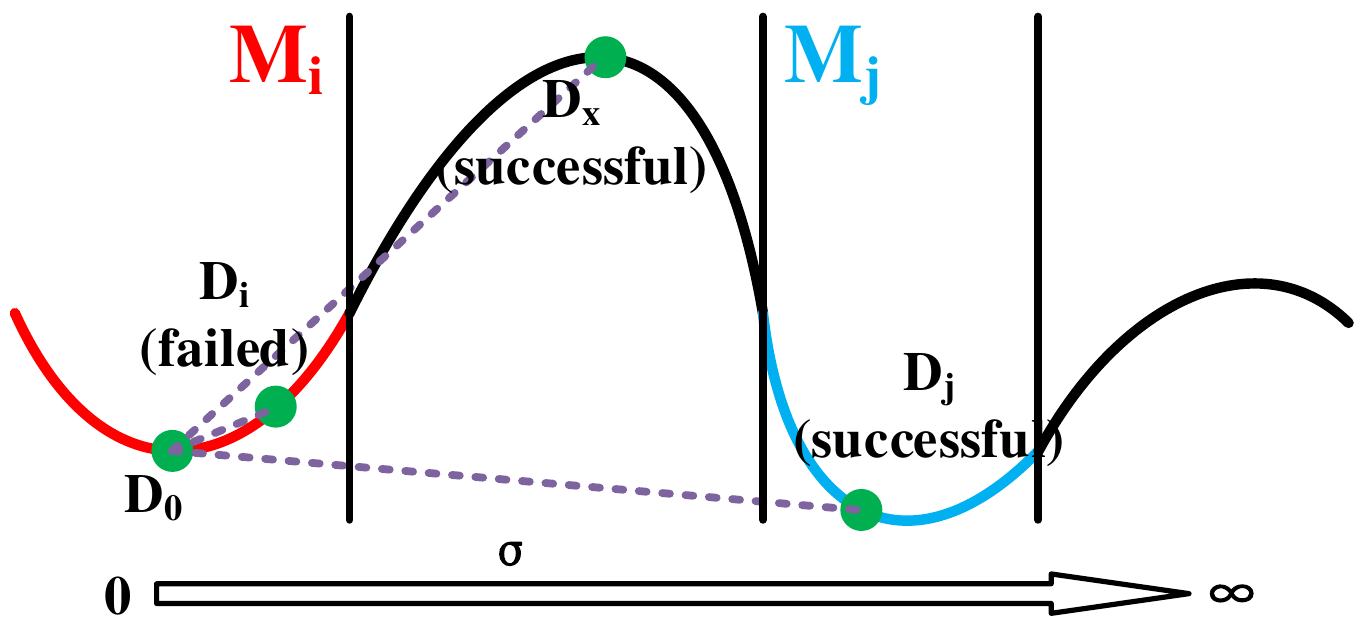}
\caption{Attack details from the manifold perspective.}
\label{manifold_2d}
\end{figure}

A relatively larger $\sigma$ may help the data `jump' out of the local minima,
with illustration shown in Figure~\ref{manifold_2d}.
$M_i$ and $M_j$ are two local minima of class $i$ and $j$.
A larger $\sigma$ is better at driving $D_0$ out of $M_i$.
Although $M_i$ may exist in areas with even larger $\sigma$,
such probability is much smaller than the data stuck in $M_i$ with a small $\sigma$.
A good initialization of $\sigma$ is even more important than the iteration process.
Therefore, in the experiments, we choose 10 as the initial std value,
and results in Section \ref{differentscale} verify our claim.

\begin{figure}[t]
\centering
\includegraphics[width=0.7\linewidth]{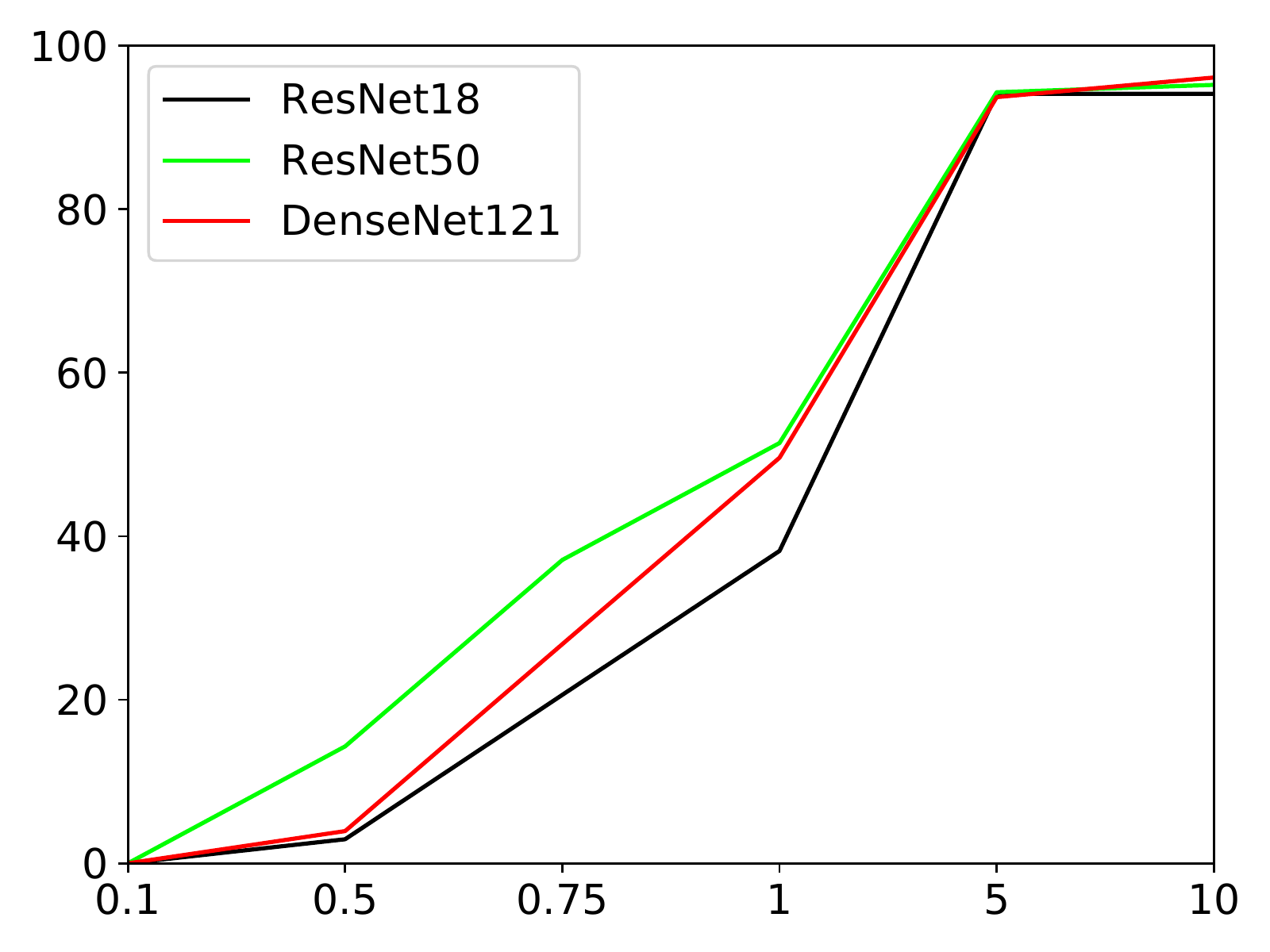}
\caption{Error rates of the three models caused by different $\sigma$ initialization in the gaussian blurring based attack.}
\label{sigma}
\end{figure}

In order to further prove the significance of $\sigma$ initialization,
we conduct experiments on the relationship between attack performance and the initial $\sigma$ value.
In the experiments, three networks structures are adopted with gaussian kernel scale 9.
With the initial value set 0.1, 0.5, 0.75, 1, 5 and 10,
the gaussian blurring based attack on the three models have the error rates shown in Figure~\ref{sigma}.
Smaller initial $\sigma$ values greatly reduce the attack performance of the method.
In contrast, iterations of $\sigma$ are much less important.

In real scenarios, if any of the three target networks serves as the image processing module in the system,
gaussian blurring with kernel scale 9 and $\sigma=10$ is obviously a wrong choice
that turns over 95\% of the ordinary data into adversarial examples.
Also according to the experiments, a larger kernel scale and a larger $\sigma$
both increase the denoising performance of image preprocessing,
but in the meantime, both are more likely to turn the ordinary data into adversarial examples.
Therefore, the trade-off between denoising performance and robustness of blurring in image preprocessing
is a significant and urgent problem.

\section{Conclusion}\label{conclusion}

In this paper, we argue that the commonly-used gaussian blurring technique
is a potentially unstable factor which may be vulnerable to well-designed adversarial attacks.
We first propose a simple demo of feature blurring for adversarial attacks.
Then, we apply gaussian blurring to the original data,
to prove that gaussian blurring can be turned aggressive in the pixel-level image preprocessing applications.
Experiment results show that the blurring demo and the gaussian blurring based attack
successfully confuse the target network and mislead it to wrong judgements.
We demonstrate from a specific perspective that an ordinary module in perception systems
may cause threats to accurate scene recognitions.

\bibliographystyle{unsrt}
\bibliography{egbib}

\end{document}